\documentclass[10pt,twocolumn,letterpaper]{article}

\usepackage{cvpr}              

\usepackage{graphicx}
\usepackage{amsmath}
\usepackage{amssymb}
\usepackage{booktabs}
\usepackage{multirow}
\usepackage{tabularray}
\UseTblrLibrary{booktabs}
\usepackage{float}
\usepackage[utf8]{inputenc}
\usepackage{enumitem}
\usepackage{listings}

\definecolor{cvprblue}{rgb}{0.21,0.49,0.74}
\usepackage[pagebackref,breaklinks,colorlinks,allcolors=cvprblue]{hyperref}

\def\paperID{13} 
\def\confName{CVPR}
\def\confYear{2025}

\title{PS4PRO: Pixel-to-pixel Supervision for Photorealistic \\Rendering and Optimization}

\vspace{-0.2cm}
\author{Yezhi Shen$^*$, Qiuchen Zhai$^*$, Fengqing Zhu\\
School of Electrical and Computer Engineering, Purdue University\\
{\tt\small \{ shen397, qzhai, zhu0 \}@purdue.edu}
}

\begin{document}
\twocolumn[{%
\renewcommand\twocolumn[1][]{#1}%

\vspace{-1cm}

\maketitle
\begin{center}
    \centering
    \captionsetup{type=figure}
    \vspace{-0.8cm}
    \includegraphics[width=.95\textwidth]{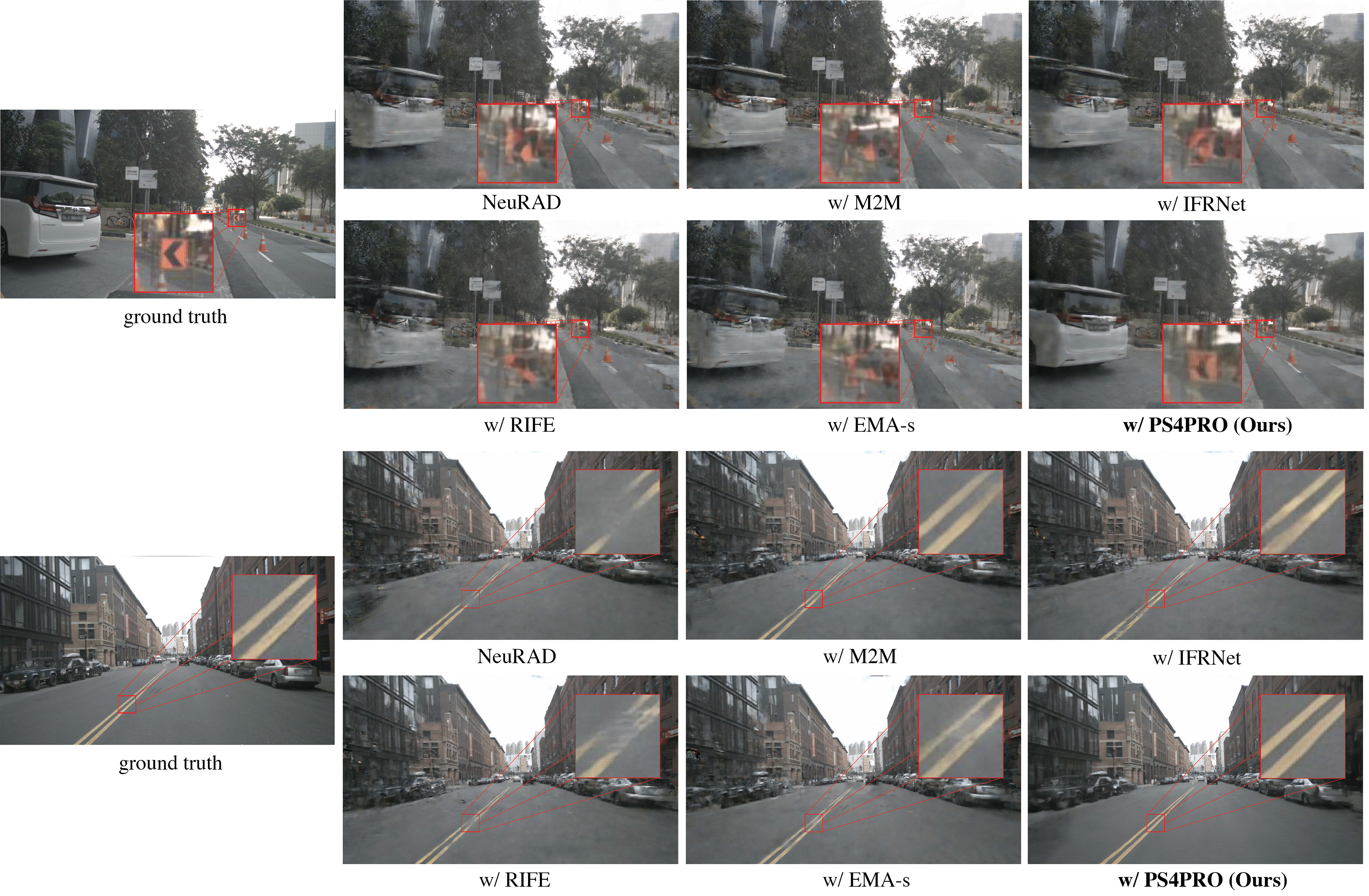}\vspace{-0.2cm}
    \caption{Reconstruction results from NeuRAD with different VFI enhancements on the NuScene-mini dataset. Our proposed method outperforms others in terms of scene clarity and completeness. Areas of interest are highlighted with red bounding boxes and zoomed in.}
    \label{fig:nuscene}
\end{center}%
}]

\renewcommand{\thefootnote}{\fnsymbol{footnote}}
\footnotetext[1]{These authors contributed equally.}

\begin{abstract}
Neural rendering methods have gained significant attention for their ability to reconstruct 3D scenes from 2D images. The core idea is to take multiple views as input and optimize the reconstructed scene by minimizing the uncertainty in geometry and appearance across the views. However, the reconstruction quality is limited by the number of input views. This limitation is further pronounced in complex and dynamic scenes, where certain angles of objects are never seen. In this paper, we propose to use video frame interpolation as the data augmentation method for neural rendering. Furthermore, we design a lightweight yet high-quality video frame interpolation model, PS4PRO (Pixel-to-pixel Supervision for Photorealistic Rendering and Optimization). PS4PRO is trained on diverse video datasets, implicitly modeling camera movement as well as real-world 3D geometry. Our model performs as an implicit world prior, enriching the photo supervision for 3D reconstruction. By leveraging the proposed method, we effectively augment existing datasets for neural rendering methods. Our experimental results indicate that our method improves the reconstruction performance on both static and dynamic scenes. 

\end{abstract}
    \vspace{-0.5cm}
\section{Introduction}
\label{sec:intro}

\begin{figure*}[h!]
    \centering
    \vspace{-0.3cm}
    \includegraphics[width=0.95\textwidth]{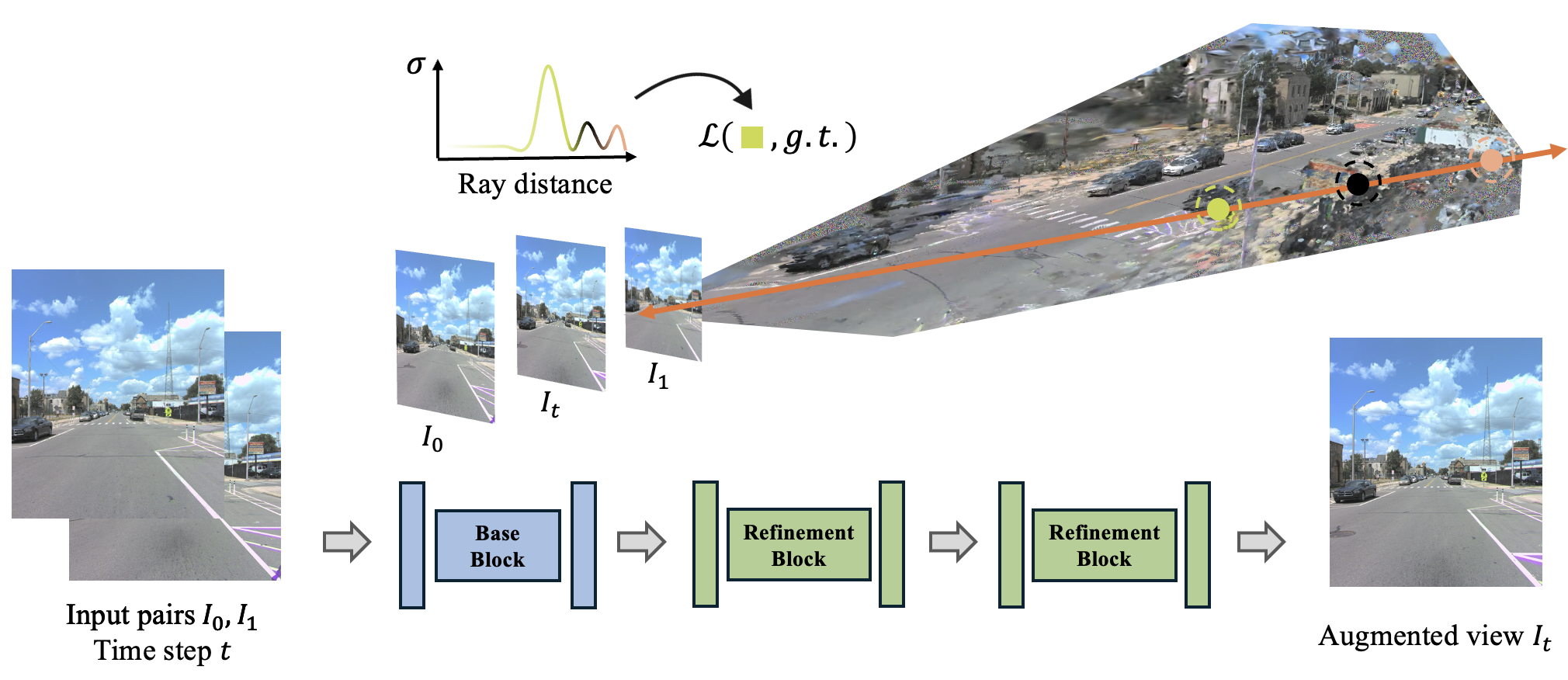}
    \vspace{-0.3cm}
    \caption{Illustration of the overall framework of the proposed method. Given input frames $I_0$ and $I_1$, the proposed frame interpolation model synthesizes the intermediate view $I_t$ at time $t$. Then both the original views and intermediate views are used to optimize the neural rendering model. Notably, the figure illustrates one interpolated frame, while the model is capable of generating multiple intermediate frames at different time steps. See Figure~\ref{fig:vfi_block} for more details on the Base Block and Refinement Block.}
    \label{fig: overall}
    \vspace{-0.3cm}
\end{figure*}

Neural Radiance Fields (NeRF)  ~\cite{nerf} and 3D Gaussian Splatting (3DGS)~\cite{3dgs} are advanced neural rendering methods based on deep learning. These methods are developed to generate high-quality, realistic 3D scenes from 2D images. By achieving high-fidelity scene reconstruction, these methods enable the rendering of sensor-accurate data from novel viewpoints, making them valuable for tasks such as view synthesis, 3D modeling, and pose and shape estimation~\cite{ingp, mip}. Considering the fundamental similarities between NeRF and 3DGS methods in terms of input and neural rendering processes~\cite{chen2024survey, wu2024recent}, we refer to both as neural rendering in the following discussion for the sake of clarity and simplicity. 

As neural rendering techniques have evolved, they are increasingly used to create detailed 3D representations to enhance perception in real-world applications, such as reconstruction for autonomous driving scenes~\cite{ming2024benchmarking}. In such dynamic scenes, objects are continuously moving and interacting within the scene, making it more challenging for volumetric functions to understand the scene and update the representation of the objects~\cite{rabich2024fpo++}. Accurate reconstruction of dynamic scenes requires extensive scene coverage~\cite{unisim, neurad, mip360}. Incomplete and insufficient coverage often results in inaccuracies and artifacts, leading to gaps in geometric information and inconsistencies in scene representation~\cite{ma2023deformable, yu2024viewcrafter, 3dgsEh}.

To address these limitations, methods have explored the use of multiview encoders~\cite{pixelnerf, gu2024egolifter}, monocular depth estimation~\cite{cf3dgs}, and diffusion model~\cite{yu2024viewcrafter} to enhance the fidelity of the reconstruction. Multiview encoder estimates view correspondence from input views and perform the reconstruction. However, the multiview-based approach relies on prior knowledge that typically requires pretraining on specific categories of scenes, which limits the generalization ability of these models across diverse scenarios. The monocular depth estimation predicts from each single image but often struggles with accuracy, especially in complex scenes. Although diffusion models can handle complex scenes, the training and inference are computationally intensive and introduce inconsistent details in the predictions~\cite{chang2024how}. Video frame interpolation (VFI) is used to synthesize intermediate frames between existing ones~\cite{parihar2022comprehensive} and has shown promising results in many video processing applications. However, to the best of our knowledge, its potential in the context of static and dynamic scene reconstruction remains unexplored.

Since VFI models are trained on large-scale video datasets, they contain implicit awareness of the scene structures as well as the camera motions \cite{guo2024free}. In this paper, we propose to leverage the knowledge of VFI models to supervise multiview consistency in 3D reconstruction, attain pixel correspondence information, and achieve little time penalty during scene optimization. We introduce a lightweight, flow-based video frame interpolation model, PS4PRO (Pixel-
to-pixel Supervision for Photorealistic Rendering and Optimization), designed to enhance neural rendering reconstruction by interpolating input images. Our model generates novel intermediate views to provide essential pixel correspondence from sparser training data. By effectively interpolating neighboring frames, our method minimizes the uncertainty of pixel correlation in 3D space, thereby improving the model's ability to infer accurate scene representation. We demonstrate PS4PRO's generalizability across multiple neural rendering methods and scenarios. The results indicate that our method consistently achieves notable improvements across tested implementations and diverse real-world scenarios. Our key contributions are summarized as follows:
\setlist{nolistsep}
\begin{itemize}[itemsep=0.1em]
 
    \item Our method introduces pixel-to-pixel supervision on neural rendering of dynamic scenes, extending the capability of neural rendering methods;

    \item We propose a simple yet effective method for imposing pixel-to-pixel supervising to neural rendering by performing real-time frame interpolation on the training frames, which has negligible computation cost;

    \item We design a lightweight frame interpolation model that can comprehend 3D geometry knowledge from training data, leading to better 3D reconstruction quality when used for neural rendering data augmentation;
    \item Extensive experiments demonstrate that our method generalizes well on different datasets without the need for fine-tuning on specific recordings.
\end{itemize}

\section{Related Works}
\label{sec:related}

\textbf{Video Frame Interpolation:} Video frame interpolation (VFI) technique has been widely studied in recent years due to its significance in various video processing applications, including generating smooth slow-motion videos, increasing video frame rates, and enhancing visual quality. 
Conventional VFI methods rely on model-based motion estimation, blending, and morphing techniques~\cite{choi2007motion, parihar2022comprehensive}, which can be computationally intensive and prone to artifacts such as ghosting or blurring.
Recent advances in deep learning have significantly improved VFI techniques, addressing many limitations of conventional methods. 
This has led to the emergence of end-to-end neural networks for VFI, such as Depth-Aware Video Frame Interpolation (DAIN)~\cite{DAIN}, Real-Time Intermediate Flow Estimation (RIFE)~\cite{RIFE}, Many-to-Many Splatting (M2M)~\cite{m2m}, and EMA-VFI~\cite{EMA}. 
These methods explore various approaches including mixed, sequential, and parallel feature extractions, and leverage advanced techniques like deformable convolution and depth estimation to produce more accurate and visually appealing interpolated frames. 
Despite the improvements, challenges such as handling large motion artifacts and preserving fine details remain, driving further research in this area.

\textbf{Autonomous Driving NeRF:} Autonomous driving researchers have incorporated neural radiance fields (NeRF) to reconstruct 3D scenes and simulate safety-critical scenarios. UniSim~\cite{unisim} introduces a unified simulation approach that leverages NeRF-based scene representations to generate highly realistic synthetic data, significantly aiding the training of autonomous driving models. Neurad~\cite{neurad} pushes this further by optimizing NeRF for autonomous driving, focusing on modeling physical sensors and improving the fidelity of object-level details to shorten the simulation to real-world gap. Meanwhile, Lightning-NeRF~\cite{cao2024lightning} addresses the computational challenges of traditional NeRF implementations by introducing fast neural initialization techniques, allowing for efficient and scalable 3D scene reconstructions. These methods collectively advance the integration of NeRFs in autonomous driving by improving both the quality and speed of 3D scene representations.

\textbf{Neural Rendering Enhancement:} To address challenges in neural rendering arising from insufficient or low-quality data, several methods have been developed. AlignNeRF~\cite{jiang2023alignerf} introduces an optical-flow network to enhance view alignment during training, thereby improving high-frequency details in reconstructed scenes. DiffusionNeRF~\cite{wynn2023diffusionerf} employs a diffusion model to learn gradients of RGBD patch priors, providing regularized geometry and color information for each scene. The 3DGS-enhancer~\cite{3dgsEh} incorporates the diffusion process after initial reconstruction to refine the quality of novel views. However, while effective on low-quality reconstructions, these methods introduce artifacts to high-quality reconstructions and substantially extend the training time, further prolonging an already lengthy process~\cite{yu2024viewcrafter}.
\section{Method}
\label{sec:method}

In this section, we describe our proposed method, which combines VFI models with neural rendering techniques to enhance 3D reconstruction.

\subsection{Neural rendering}
Neural Radiance Fields (NeRF) and 3D Gaussian Splatting (3DGS) represent two distinct yet conceptually unified approaches in neural rendering. Both methods fundamentally rely on an optimization framework where the primary objective is to reconstruct a 3D scene by minimizing discrepancies between the rendered outputs and a set of ground truth images. At their core, both NeRF and 3DGS use differentiable rendering techniques to learn the 3D scene structure by casting rays from the camera into the scene, estimating color and opacity along these rays, and aggregating the results to generate 2D projections. The only difference between the two lies in the representation of the underlying 3D scene, with NeRF relying on implicit neural representations and 3DGS using a more explicit data-driven approach with point-based splatting.

The optimization in both NeRF and 3DGS can be formalized as a minimization problem, where the goal is to adjust the model parameters such that the rendered image $\hat{I}$ is as close as possible to the observed image $I$. Given a set of camera positions, for each pixel $p$ in the image, a ray $r(t) = \boldsymbol{o}+t\boldsymbol{d}$ is cast from the camera origin $\boldsymbol{o}$ in the direction $\boldsymbol{d}$. 
 
The optimization objective can be expressed as:
\begin{equation}
    \mathcal{L} = \sum_{p} \left\| I(p) - \hat{I}(p) \right\|^2 ,
\end{equation}
where \(I(p)\) is the observed color value at pixel \(p\) and \(\hat{I}(p)\) is the model's predicted color for that pixel. This loss function \(\mathcal{L}\) drives the adjustment of parameters to better match the rendered outputs with the ground truth.

The rendering process in both NeRF and 3DGS involves aggregating the estimated color and opacity along a ray. For a given ray \(r(t)\), the color is computed by integrating the contributions of the scene along the ray's path using the formula:
\[
C(r) = \int_{t_{n}}^{t_{f}} T(t) \sigma(r(t)) c(r(t)) \, dt ,
\]
where \(t_{n}\) and \(t_{f}\) are the near and far bounds of the integration, \(T(t) = \exp\left( -\int_{t_{n}}^{t} \sigma(r(s)) \, ds \right)\) is the accumulated transmittance representing the probability that light travels without being occluded until distance \(t\), \(\sigma(r(t))\) is the volume density at position \(r(t)\), and \(c(r(t))\) is the color contribution. This formulation ensures that both methods aggregate information from the entire ray path to produce realistic images that match the input data as closely as possible.

\subsection{Pixel-to-pixel supervision}

\begin{figure}[h!]
    \centering
    \includegraphics[width=0.4\textwidth]{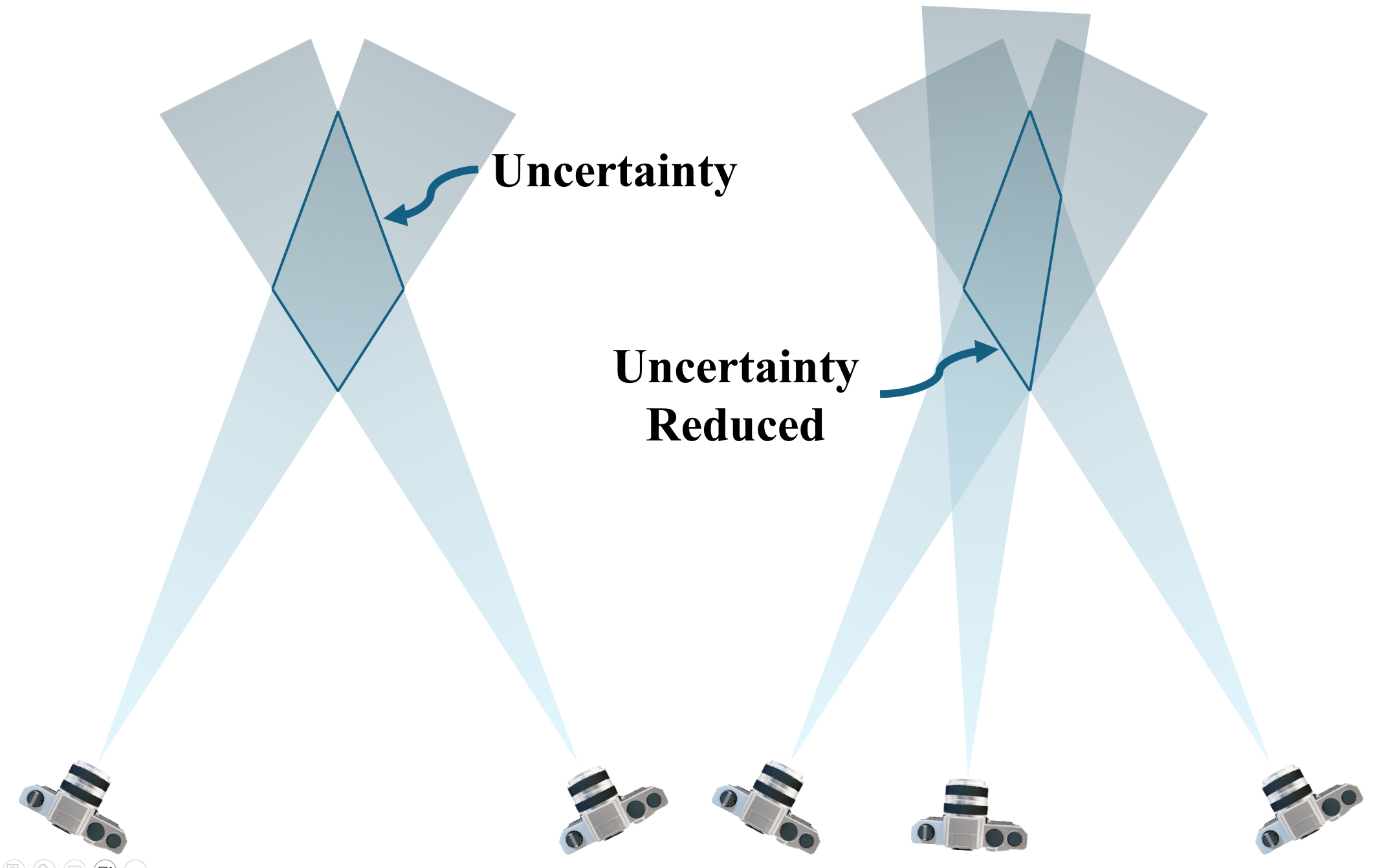}
    \caption{Illustration of reduction of the reconstruction uncertainty by introducing intermediate frames as supervision.}
    \label{fig:p2p}
\end{figure}

Pixel-to-pixel supervision is crucial for better motion understanding and 3D scene reconstruction. Assuming the substance observed by each pixel is non-transparent and self-emitting~\cite{3dgs, fan2023lightgaussian}, enforcing consistent predictions across multiple views allows for more accurate correspondence between pixels, thus reducing the matching ambiguity and enhancing the smoothness of estimates~\cite{mvsnerf}. Such supervision ensures that small discrepancies in predictions are corrected in overlapping ray regions, leading to more accurate reconstruction of scenes. 

In the context of neural rendering methods, pixel-to-pixel supervision can be employed by optimizing consistency in predicted color and density along different directions and matching pixels between frames. This typically involves establishing the correspondence between pixels across two frames. Let $p$ represent the pixel position of a substance in the first frame $I_0$, and $p'$ denote that in the second frame $I_1$.
The relation between the corresponding pixel positions $p$ and $p'$ can be depicted by 
\begin{equation}
    p' = p + q ,
\end{equation}
where $q$ is the displacement vector between these pixels. 
To find the correspondence between frames, the flow vector $q$ can be estimated by analyzing the pixel motion or using advanced neural networks. 

In our work, we assess the pixel correspondence between consecutive frames $I_0$ and $I_1$ externally and use this information to facilitate neural scene reconstruction. This is achieved by generating novel views between the frames based on the following pixel correspondence.
\begin{equation}
    I_0(p) = I_{1} (p') ,
\end{equation}
which minimizes the uncertainty in geometry arising from intersecting cross-sections of rays.

As the neural rendering methods face uncertainty when aggregating information from intersecting rays relying on sparse supervision, pixel-to-pixel supervision can further reduce the ambiguity in cases where multiple rays overlap in their cross-sections.

As illustrated in Figure~\ref{fig:p2p}, by anchoring the contribution of each ray, the pixel-to-pixel supervision reduces ambiguity by mitigating interference from intersection areas.

\subsection{VFI model}
\label{sec:vfi}

To improve the performance of scene reconstruction methods, we propose our video frame interpolation model, PS4PRO, based on pixel-to-pixel supervision, which helps reduce pixel uncertainty by supervising the reconstruction at the pixel level. In addition, we leverage the benefits of world knowledge by training on diverse scenes and datasets. The traditional pixel correspondence supervision approaches are typically limited by the homogeneity of the training data, while our integration of rich and diverse scene information enhances the model's ability to generalize, leading to more accurate and precise reconstructions across different types of scenes.

Our goal of the video frame interpolation model is to generate novel view $I_t \in \mathbb{R}^{H\times W \times 3}$ at an arbitrarily given timestep $t$ from the existing left and right frame $I_0$, $I_1 \in \mathbb{R}^{H\times W \times 3}$. Previous works~\cite{jiang2018super, abme, bmbc} have modeled the problem of frame interpolation as finding the linear intermediate of $I_0$ and $I_1$. This approach is incomplete for end-to-end trained VFI models, as they capture knowledge of the 3D geometries from the training data. Thus, our PS4PRO model can be represented by
\begin{align}
  I_t &= \mathbb{O}(I_0, I_1, t) | K_{w} ,
\end{align}
where $\mathbb{O}$ denotes the process of frame interpolation, and $K_w$ represents the learned world prior consisting of camera motion and 3D geometry of captured scenes.

The proposed video frame interpolation model handles appearance and motion information in a mixed way, where the two neighboring frames are directly concatenated and fed into a backbone to generate motion representation. A novel view is generated by combining the warped left and right images according to the predicted motion maps $F_{t\rightarrow 0}, F_{t\rightarrow 1}$ and mask $M$,
\begin{equation}
  \label{eqn:merge}
  I_t = \overleftarrow{w}(I_0, F_{t\rightarrow 0})*M + \overleftarrow{w}(I_1, F_{t\rightarrow 1})*(1-M) ,
\end{equation}
where the $\overleftarrow{w}$ denotes the backward warping operation.

The flow estimation architecture of our model is illustrated in Figure~\ref{fig: overall}, where the input frames $I_0$ and $I_1$ go through three layers of feature pyramids, a base block, and two refinement blocks. Designed to advance from accuracy to precision~\cite{mmsp}, the base block predicts global motion vectors at low resolution, and the following refinement blocks perform motion refinement at increasing resolutions. Each of the blocks consists of two hourglass-shape modules with weights shared to extract $F_{t\rightarrow 0}$ and $ F_{t\rightarrow 1}$.

 \begin{figure}[b!]
    \centering
    \includegraphics[width=0.4\textwidth]{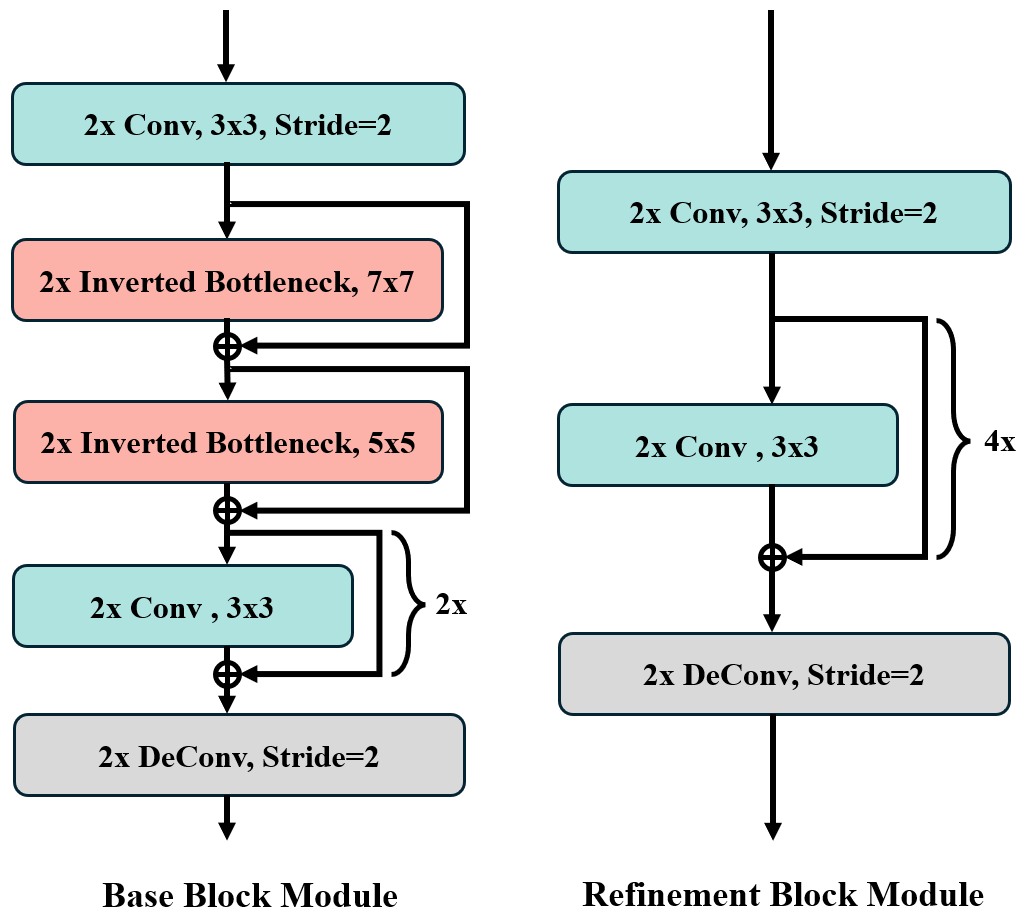}
    \caption{Architecture of Base and Refinement Blocks in PS4PRO.}
    \label{fig:vfi_block}
\end{figure}

\textbf{Base Block:} As shown in Figure~\ref{fig:vfi_block}, the base block module is structured with an hourglass architecture, which is particularly advantageous for spatial compression while retaining detailed motion estimation at the input resolution. In PS4PRO, the base block maps coarse motion vectors with high accuracy by maximizing the receptive field. This block is composed of six convolutional groups, each containing two layers, to handle hierarchical feature extraction. The initial stem group consists of two layers of $3\times3$ convolutions with a stride of 2, achieving down-sampling and pixel aggregation with learnable parameters. Following the stem, two groups employ inverted bottleneck structures with an expansion rate of 2, as outlined in~\cite{convnext}. These groups are optimized for capturing broad-range features and performing channel lifting. Specifically, the first inverted bottleneck group has a kernel size of $7\times7$, while the second utilizes a $5\times5$ kernel size to provide a multi-scale feature extraction. The fourth and fifth groups are designed with $3\times3$ convolutions, tailored for finer feature extraction while maintaining computational efficiency. The output block concludes with two de-convolution layers that up-sample the features back to the original resolution, aligning them with the output dimensions. Additionally, skip connections are applied across each convolutional layer group, ensuring that downstream layers retain access to the initial image information, promoting enhanced reconstruction accuracy.

\textbf{Refinement Block:} The refinement block module operates under the assumption that the base block has already provided an accurate initial prediction of the motion between input frames. Its primary role is to up-sample these predicted motion vectors from the base layer and refine them at higher resolutions for finer details. Consequently, the refinement block does not require an extensive receptive field, allowing it to concentrate on local pattern refinement. The design of the refinement block mirrors that of the base block, though it incorporates several key modifications to optimize its function. Specifically, the two groups of inverted bottleneck layers in the base block are replaced with two groups of $3\times3$ convolution layers, which are better suited for localized adjustments and further enhance the refinement process.

\textbf{Step Control:} Many previous works engineered sophisticated backward flow reversal layers to handle arbitrary step flow estimation~\cite{bmbc, abme, EMA}, resulting in flaws in motion boundaries due to the high complexity of optical flows. RIFE proposed an end-to-end method for time step incorporation by adding a layer of time step as input at every stage of the flow estimation network. We argue that since the base block has already provided an accurate but less precise prediction, the refinement blocks do not need the time step as auxiliary information, as the corresponding pixels in the warped left and right image have a similar magnitude of motion prediction error in opposite directions. Adding the time step to the input of the refinement blocks confuses the flow estimation process. In the design of PS4PRO, the time step is only passed into the base block.
\section{Experiments}
\label{sec:experiments}

\subsection{Dataset}

\begin{table*}[t]
  \vspace{-0.3cm}
  \centering
  \begin{tabular}{l c c c c c c c c c c c c c}
    \toprule
    \multirow{2}{3em}{Method} & \multirow{2}{2.5em}{Vimeo} & \multirow{2}{2em}{UCF} & \multirow{2}{2em}{M.B.} & \multicolumn{4}{c}{SNU-FILM} & \multicolumn{3}{c}{HD} & \multicolumn{2}{|c}{X4K1000FPS} & \multirow{2}{2em}{Param.}\\
    \cmidrule{5-13}
    & & & & Easy & Med. & Hard & Extr. & 544p & 720p & 1080p & 2K & 4K & \\
    \midrule
    M2M & 35.47 & 35.28 & \underline{2.09} & \underline{39.65} & \textbf{35.71} & 30.28 & 25.08 & 22.31 & \underline{31.94} & 33.45 & \textbf{32.13} & \underline{30.88} & 7.61M\\
    IFRNet & 32.36 & 33.59 & 3.30 & 37.34 & 33.74 & 29.35 & 24.70 & 22.01 & 31.85 & 33.19 & 31.53 & 30.46 &  4.96M\\
    RIFE & 34.84 & 35.22 & 2.28 &  39.58 & 35.53 & 30.45 & 25.57 & 22.95 & 31.87 & 34.25 & 31.43 & 30.58 & 10.7M\\
    EMA-s & \underline{35.48} &  \textbf{35.34} & 2.18 & 39.60 & 35.55 & \textbf{30.71} & \textbf{26.02} & \textbf{23.26} & \textbf{32.17} & \textbf{34.65} & \underline{31.89} & \textbf{30.89} & 14.3M\\
    Ours & \textbf{35.64} & \underline{35.32} & \textbf{2.07} & \textbf{39.71} & \textbf{35.79} & \underline{30.53} & \underline{25.78} & \underline{23.02} & 31.90 & \underline{34.33} & 31.86 & 30.79 & 5.09M\\
    \bottomrule
  \end{tabular}
  \caption{Quantitative comparison of arbitrary timestep video frame interpolation methods on different datasets (IE on Middlebury, PSNR on other datasets) and model parameter size. Our model achieves the best performance in center frame interpolation tasks.}
  \label{tab:fixed}
  \vspace{-0.3cm}
\end{table*}

In this paper, we utilize nine datasets for training and evaluation, each providing its distinct characteristics in video processing and scene reconstruction. 
\textbf{1) Vimeo90K}~\cite{xue2019video} consists of two subsets, Triplet and Septuplet, and contains video sequences at a fixed resolution of $448 \times 256$. The Triplet subset includes sequences with three consecutive video frames, while each video sequence in the Septuplet subset contains seven frames.
\textbf{2) UCF101}~\cite{soomro2012ucf101} contains 379 video triplets capturing various human motions, with each sample at a resolution of $256 \times 256$. 
\textbf{3) MiddleBury (M.B.)}~\cite{baker2011database} includes image pairs with a resolution of $640 \times 480$. In this paper, we utilize the 'OTHER' subset due to its realistic complexity and diversity.
\textbf{4) SNU-FILM}~\cite{choi2020channel} contains four subsets-'Easy', 'Medium', 'Hard', and 'Extreme', with a total of 1,240 triplets at a resolution of $1280 \times 720$. 
\textbf{5) HD}~\cite{bao2019memc} is composed of eleven videos at three different resolutions, 544p, 720p, and 1080p, which allows evaluation across varying scales. We follow the evaluation outlined in~\cite{RIFE} for evaluation consistency. 
\textbf{6) X4K1000FPS}~\cite{xvfi} consists of 4K video samples captured at a high frame rate. We follow~\cite{xvfi} for evaluation on 2K and 4K resolutions. 
\textbf{7) KITTI}~\cite{liao2022kitti} provides traffic scenarios with various sensor modalities, including images at a resolution of $1408 \times 376$ and lidar scans. We follow~\cite{lu2023urban, cao2024lightning} for evaluation on this dataset. 
\textbf{8) Argoverse2} (Argo)~\cite{argoverse} comprises images at a resolution of $1550 \times 2048$ and lidar scans. We align our evaluation on this dataset with the methods described in~\cite{cao2024lightning}. 
\textbf{9) NuScenes-mini}~\cite{nuscenes} is a subset of the full NuScenes autonomous driving dataset, consisting of images at a resolution of $1600 \times 900$ along with Lidar scans.

Utilizing these datasets collectively enables a comprehensive assessment of our method, as each dataset introduces unique challenges and perspectives across diverse scenes. 

\subsection{Training}
We utilize the two subsets of Vimeo90K to train our model under different interpolation conditions. We train our model on a mixed fixed timestep and arbitrary timestep interpolation interleavingly. Training on the Triplet subset allows the model to predict the center middle frame more accurately. For the arbitrary timestep interpolation, we leverage the Septuplet subset, where three frames are randomly selected in order from each clip, and the timesteps are calculated accordingly. We sort the sampled three frames for 80\% of chance and we train the model for extrapolation for the remaining 20\%. This enables our model to better perceive the structure of the scenes and generalize to unseen datasets.

VFI models predicting backward optical flow do not explicitly represent the bi-directional flow between the input image pairs, making it harder to converge on complex datasets. To mitigate this problem, we incorporate a pre-trained forward optical flow VFI model following \cite{jeong2024ocai} as our teacher model.

We apply a random crop of $256 \times 256$ to each image triplet sample and perform random flipping, time reversal, rotation, and channel permutation for data augmentation. The training batch size is set to 16 per card. We choose AdamW~\cite{adamw} as the optimizer, with $\beta_1 = 0.9$ and $\beta_2 = 0.999$. The weight decay parameter is set to $1e-4$. We apply the combination of $l1$ loss and perceptual loss against ground truth ($\hat{I_t}$) to the synthesized image ($I_t$) as supervision. In the first 200 epochs, we apply the distill the backward optical flow from the teacher model using $\mathcal{L}_{tea}$.
\begin{align}
    \mathcal{L}_1 &= ||\hat{I_t}, I_t||_1 \ , \\
    \mathcal{L}_{tea} &= ||\hat{F_t}, F_t||_1 \ , \\
    \mathcal{L}_{perceptual} &= ||\Phi(\hat{I_t}), \Phi(I_t)||_2 \ , \\
    \mathcal{L} &= \mathcal{L}_1 + \lambda \times \mathcal{L}_{perceptual} +  \mathcal{L}_{tea},
    \label{eqn_loss}
\end{align}
where $\hat{F_t}$ is the backward optical flow form the teacher model, $F_t$ is the predicted flow, and the weighting parameter $\lambda$ is set to 0.005. 

We trained our model on four NVIDIA A5000. During training, we first warm up for 2,000 steps while gradually increasing the learning rate to 3e-4. Then we utilize cosine annealing to reduce the learning rate from 3e-4 to 3e-6. We train our method for 300 epochs on the training datasets and compare our model with the recent advanced works including RIFE~\cite{RIFE}, M2M~\cite{m2m}, IFRNet~\cite{kong2022ifrnet}, and EMA-s~\cite{EMA}. In all sections below, we mark the best result in \textbf{Bold} and the second-best result in \underline{underline} in each comparison table. 

\subsection{Evaluation on Frame Interpolation Datasets}
To demonstrate the effectiveness of our method, we conduct experiments on multiple datasets, including fixed timestep and arbitrary timestep datasets, comparing the performance of PS4PRO against other arbitrary timestep interpolation models. For the fixed timestep datasets, we evaluated the VFI methods on the datasets including Vimeo90K, UCF101, Middlebury, and SNU-FILM. For the arbitrary timestep video frame interpolation datasets, we evaluate the performance of the video frame interpolation methods on HD and X4K1000FPS. Table~\ref{tab:fixed} shows the quantitative comparison of the VFI methods evaluated using peak signal-to-noise ratio (PSNR). Our model achieves the best performance on Vimeo90k Triplet, Middle Burry, and two subsets of SNU-FILM. Our model also achieves the second highest scores on UCF101, two subsets of SNU-FILM and HD.

\subsection{Time Proportion of VFI Augmentation}
To demonstrate the minimal computational cost of the proposed data augmentation framework for neural scene reconstruction, we report the time required for data augmentation on each dataset, along with its proportion relative to the total neural scene reconstruction training time, as shown in Table~\ref{tab:proportion}. Details of the neural rendering experimental setup are provided in Section \ref{sec::neurand}. The data augmentation time reflects the duration each VFI method takes to augment the entire dataset. For the time of reconstruction, we averaged the training times across all scenes in each dataset over five runs, using this average as the representative training time. The results indicate that the time added by using VFI for data augmentation is constant regardless of the 3D scene optimization time and negligible compared to the overall duration of neural scene reconstruction.

\begin{table}[h]
  \centering
  \setlength{\tabcolsep}{4pt}
  \begin{tabular}{l c c c c c c}
    \toprule
    \multirow{2}{3.5em}{Methods} & \multicolumn{3}{c}{Augmentation Time $\;$} & \multicolumn{3}{|c}{Proportion of Time}\\
    \cmidrule{2-7}
    & KIT. & Arg. & NuS. & KIT. & Arg. & NuS.\\
    \midrule
    M2M & 11.5s & 174s & 120s & 0.85\text{\textperthousand} & 7.69\text{\textperthousand} & 0.47\text{\textperthousand} \\
    IFRNet & 8.09s & 163s & 100s & 0.60\text{\textperthousand} & 7.22\text{\textperthousand} & 0.39\text{\textperthousand} \\
    RIFE & \textbf{6.06}s & \textbf{111}s & \textbf{74.1}s & \textbf{0.45}\text{\textperthousand} & \textbf{4.94}\text{\textperthousand} & \textbf{0.29}\text{\textperthousand} \\
    EMA-s & 11.6s &	224s & 146s & 0.85\text{\textperthousand} & 9.91\text{\textperthousand} & 0.57\text{\textperthousand} \\
    Ours & \underline{7.69}s & \underline{155}s & \underline{96.6}s & \underline{0.57}\text{\textperthousand} & \underline{6.87}\text{\textperthousand} & \underline{0.38}\text{\textperthousand} \\
    \bottomrule
  \end{tabular}
  \caption{The total augmentation time in seconds (s) of different VFI methods on Argoverse2 (Arg.), KITTI (KIT.) and NuScenes (NuS.) datasets, and the proportion of the augmentation time in permille (\text{\textperthousand}) in the entire neural rendering process.}
  \label{tab:proportion}
  \vspace{-0.5cm}
\end{table}

\subsection{Evaluation on Neural Rendering Methods} \label{sec::neurand}

To demonstrate that video frame interpolation models can be applied to neural rendering methods to improve reconstruction quality, and our PS4PRO learns the best 3D geometry among existing VFI models, we apply the VFI methods to Lightning-Nerf~\cite{cao2024lightning} (L-NeRF) and NeuRAD~\cite{neurad} and evaluate the quality of the reconstructions. The evaluation follows the original neural rendering methods, where we report the average PSNR, SSIM, and the Learned Perceptual Image Patch Similarity (LPIPS) over each scene of each dataset.

\begin{table}[h]
  \centering
  \setlength{\tabcolsep}{4pt}
  \begin{tabular}{l c c}
    \toprule
    Method & Argoverse2 & KITTI\\
    \midrule
    L-NeRF & 30.49/0.8684/0.3512 & 23.60/0.7588/0.3488 \\
    w/ M2M & 30.51/0.8712/\underline{0.3435} & 23.74/0.7617/0.3476\\
    w/ IFRNet & 30.51/0.8706/0.3455 & \underline{23.77}/\underline{0.7620}/0.3474\\
    w/ RIFE & 30.56/\underline{0.8715}/0.3434 & 23.71/0.7609/0.3478\\
    w/ EMA-s & \underline{30.57}/\underline{0.8715}/\underline{0.3435} & 23.72/0.7615/\underline{0.3470}\\
    
    w/ Ours & \textbf{30.72}/\textbf{0.8732}/\textbf{0.3423} & \textbf{24.25}/\textbf{0.7789}/\textbf{0.3395}\\
    \bottomrule
  \end{tabular}
  \caption{Quantitative comparison (PSNR/SSIM/LPIPS) of reconstruction quality on Argoverse2 and KITTI when different VFI methods are applied to Lightening NeRF.}
  \label{tab:lnerf}
  \vspace{-0.3cm}
\end{table}

In the first experiment, we trained the L-NeRF model on the Argoverse2 dataset for 30k iterations and applied the five VFI methods for data augmentation. The Argoverse2 dataset captures scenes with dense views and bright lighting conditions, making it ideal for assessing the performance of our proposed method in controlled, stable environments. 

\begin{figure*}[t!]
  \vspace{-0.4cm}
    \centering
    \includegraphics[width=0.98\textwidth]{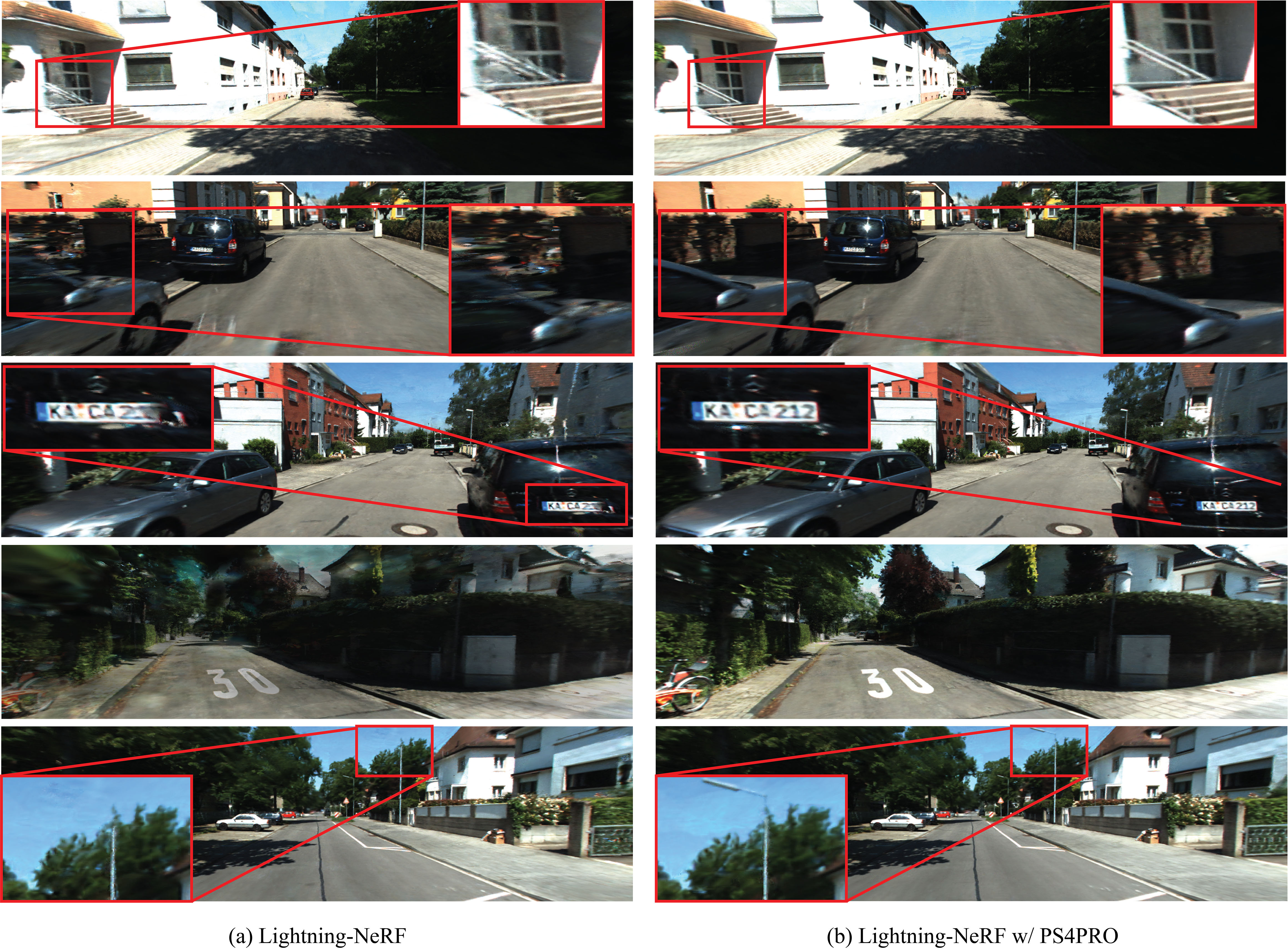}
    \caption{The visual comparison of KITTI dataset reconstructed using Lightning-NeRF without (left column) and with (right column) the data augmentation using PS4PRO. The image pairs from top to bottom are extracted from sequences 1 to 5, respectively.}
    \label{fig:kitti_result}
  \vspace{-0.5cm}
\end{figure*}

In the second experiment, we trained the L-NeRF model on the KITTI dataset for 30k iterations. KITTI dataset also focuses on static scenes but contains sparse views and introduces large contrasts with a wide field of view. Evaluating the KITTI dataset examines how different interpolation models handle more complex motions including light intensity shifts and nonuniform motions. Figure~\ref{fig:kitti_result} shows that our incorporating PS4PRO for training image augmentation significantly enhances reconstruction quality. This improvement is evident in several key areas, including better color consistency, improved geometric accuracy, and reduced noise in the rendered views.

The training set is interpolated with one extra frame in between each frame for both the Argoverse2 and KITTI dataset. Table~\ref{tab:lnerf} shows that in both experiments using VFI as data augmentation improves the reconstruction quality on static scenes, and our PS4PRO stands out by providing the most performance uplift.

\begin{table}[h]
  \centering
  \begin{tabular}{l c}
    \toprule
    Method & NuScenes-mini\\
    \midrule
    NeuRAD & 22.48/0.6601/0.3252 \\
    w/ M2M & \underline{22.69}/0.6606/\textbf{0.3199} \\
    w/ IFRNet & 22.63/\underline{0.6614}/0.3227\\
    w/ RIFE & 22.62/0.6609/\underline{0.3220}\\
    w/ EMA-s & 22.64/0.6608/\underline{0.3220}\\
    w/ Ours & \textbf{23.10}/\textbf{0.6678}/0.3239\\
    \bottomrule
  \end{tabular}
  \caption{Quantitative comparison (PSNR/SSIM/LPIPS) of reconstruction quality on NuScenes-mini when different VFI methods are applied to NeuRAD.}
  \label{tab:NeuRAD}
  \vspace{-0.2cm}
\end{table}

For the third experiment, we reconstructed the road scenes with dense traffic in NuScenes using NeuRAD, to evaluate how well the interpolation models adapt to scenes with non-static, moving traffic agents. This dataset includes moving vehicles, changing environments, and low-light scenarios, posing a greater challenge for our method. 
The NuScenes dataset lacks valid camera pose due to its height, roll, or pitch information being discarded \cite{nuscenes, neurad}. We select only the keyframes of the front camera as the training and test data since they are aligned with Lidar data and contain full 3D annotations, making them more reliable. We perform a $5\times$ interpolation for data augmentation. Table~\ref{tab:NeuRAD} shows that the result when using all VFI as the data augmentation method surpasses the baseline where no data augmentation is used. The reconstruction quality of NeuRAD w/Ours outperforms the rest, showing that our frame interpolation method comprehends 3D geometry better. Figure~\ref{fig:nuscene} shows the reconstructed scenes from the NuScenes-mini dataset. Both subfigures present the ground truth image and rendered scene reconstruction results using different VFI models as the augmentation method. As demonstrated, PS4PRO outperforms others in the experiments carried out, producing clearer and more accurate reconstructions. This is particularly evident in the reconstructed edges and fine details, such as the traffic sign at the top and the road marking at the bottom example. In contrast, others struggle with blurring and artifact distortions. 

\section{Conclusion}
\label{sec:conclusion}

In this paper, we present a practical data augmentation method that enhances neural scene reconstruction by generating high-quality intermediate frames from sparse input data using video frame interpolation. The proposed method is particularly valuable when acquiring comprehensive and sufficient training datasets is costly. To address the challenges presented by limited viewpoint and inadequate scene coverage, our model, PS4PRO, interpolates additional frames by extracting the motions in videos, establishing essential pixel correspondences between frames, and integrating the implicit world prior, which provides knowledge of the 3D structure of the scene. Extensive evaluations across neural rendering methods and diverse implementations demonstrate that our model improves reconstruction accuracy and achieves more detailed and reliable scene representation at negligible computational cost.

\newpage
{
    \small
    \bibliographystyle{ieeenat_fullname}
    \bibliography{main}
}

\end{document}